# A Deep Learning Framework for Boundary-Aware Semantic Segmentation

Tai An
University of Rochester
Rochester, USA

Weiqiang Huang
Northeastern University
Boston, USA

Da Xu
Worcester Polytechnic Institute
Worcester, USA

Qingyuan He
New York University
New York, USA

Jiacheng Hu
Tulane University
New Orleans, USA

Yujia Lou*
University of Rochester
Rochester, USA

*Abstract-As a fundamental task in computer vision, semantic segmentation is widely applied in fields such as autonomous driving, remote sensing image analysis, and medical image processing. In recent years, Transformer-based segmentation methods have demonstrated strong performance in global feature modeling. However, they still struggle with blurred target boundaries and insufficient recognition of small targets. To address these issues, this study proposes a Mask2Former-based semantic segmentation algorithm incorporating a boundary enhancement feature bridging module (BEFBM). The goal is to improve target boundary accuracy and segmentation consistency. Built upon the Mask2Former framework, this method constructs a boundary-aware feature map and introduces a feature bridging mechanism. This enables effective cross-scale feature fusion, enhancing the model's ability to focus on target boundaries. Experiments on the Cityscapes dataset demonstrate that, compared to mainstream segmentation methods, the proposed approach achieves significant improvements in metrics such as mIOU, mDICE, and mRecall. It also exhibits superior boundary retention in complex scenes. Visual analysis further confirms the model's advantages in fine-grained regions. Future research will focus on optimizing computational efficiency and exploring its potential in other high-precision segmentation tasks.*

## I. INTRODUCTION

In recent years, the rapid advancement of computer vision and deep learning has positioned semantic segmentation as a crucial component in tasks such as image understanding and scene parsing [1]. By classifying objects in images or videos at the pixel level, semantic segmentation helps machines better perceive and interpret the external world. In applications like autonomous driving, human-computer interaction, and medical image analysis, high-precision segmentation directly impacts system safety and reliability [2]. As a result, researchers have continuously explored ways to enhance segmentation accuracy and robustness, with a particular focus on precise target boundary delineation.

Among deep learning-based semantic segmentation algorithms, mainstream pixel-level convolutional neural networks (CNNs) can effectively classify and predict the macrostructure of a scene. However, they often overlook local details, especially subtle changes at object edges. These edge regions, which carry highly structured features in real-world environments, are critical for target shape recognition and semantic interpretation. When boundary information is lost or ignored, target edges become blurred, compromising overall segmentation quality. Consequently, achieving a harmonious balance between global and local feature representation, with a particular emphasis on enhancing object boundary precision, constitutes a pivotal challenge in the advancement of semantic segmentation research.

To overcome the limitations of traditional single-feature extraction methods, researchers have explored strategies for integrating boundary information into model structures. Introducing a feature bridging module allows the interactive fusion of multi-scale and multi-level semantic features. This enhances the model's ability to capture contours and refine segmentation accuracy. Boundary enhancement is not an isolated process but is deeply intertwined with the deep network's overall feature extraction and discrimination capabilities. By ensuring boundary information propagates effectively through the model, it strengthens high-level semantic decisions while enabling low-level features to learn more discriminative local details, ultimately improving segmentation performance.

Against this backdrop, next-generation segmentation architectures such as Mask2Former have demonstrated significant potential. Instead of traditional "pixel-level" prediction, Mask2Former shifts towards "instance-level" or "region-level" mask prediction, greatly improving its ability to parse complex scenes. By incorporating a boundary-enhancing feature bridging module into the Mask2Former framework, the model better captures target edge details and facilitates effective information exchange across multi-level feature representations [3]. This leads to further improvements in segmentation accuracy and robustness.

The Mask2Former-based semantic segmentation algorithm with boundary enhancement not only advances academic research in deep learning-based segmentation but also exhibits strong industrial applications. Its precise boundary recognition and robust feature fusion capabilities enable more efficient and reliable scene understanding in areas such as autonomous driving perception, refined medical image analysis, video surveillance, and human-computer interaction. This focus on edge information introduces new perspectives for future semantic segmentation developments, inspiring researchers to explore more comprehensive and generalizable segmentation architectures.

## II. RELATED WORK

Semantic segmentation has evolved from traditional methods like graph cuts, CRFs, and MRFs [4]—which capture basic spatial relationships but struggle with complex backgrounds—to deep learning approaches led by CNNs. Pioneering CNN-based frameworks such as Fully Convolutional Networks (FCN) [5], U-Net [6], and DeepLab leverage end-to-end training, skip connections, and dilated convolutions to improve pixel-level predictions. However, CNNs often struggle with long-range dependencies and produce blurred boundaries.

Recently, Transformer-based models (e.g., ViT [7], Swin Transformer [8], DETR [9], and Mask2Former) have emerged, demonstrating strong capability in modeling global context.

Yet, boundary details remain a challenge. To tackle this, boundary attention mechanisms and multi-scale feature fusion are increasingly employed, as seen in BASNet [10] and BiFusion Net. Building on these strategies, this study integrates a boundary enhancement feature bridging module into Mask2Former, aiming for sharper target delineation and more robust segmentation performance.

## III. METHOD

This study proposes a Mask2Former semantic segmentation algorithm based on the Boundary-Enhanced Feature Bridging Module (BEFBM) to improve the accuracy of target boundaries and segmentation consistency. Mask2Former uses a query mechanism for mask prediction and combines the Transformer architecture to model global information. However, the original Mask2Former does not pay enough attention to boundary information and easily produces blurred areas at the junction of targets. To solve this problem, this study introduces BEFBM in the feature extraction stage of Mask2Former, making the interaction between features of different scales more sufficient while enhancing the focus on target boundaries, thereby improving the overall segmentation effect. The model architecture of Mask2Former is shown in Figure 1.

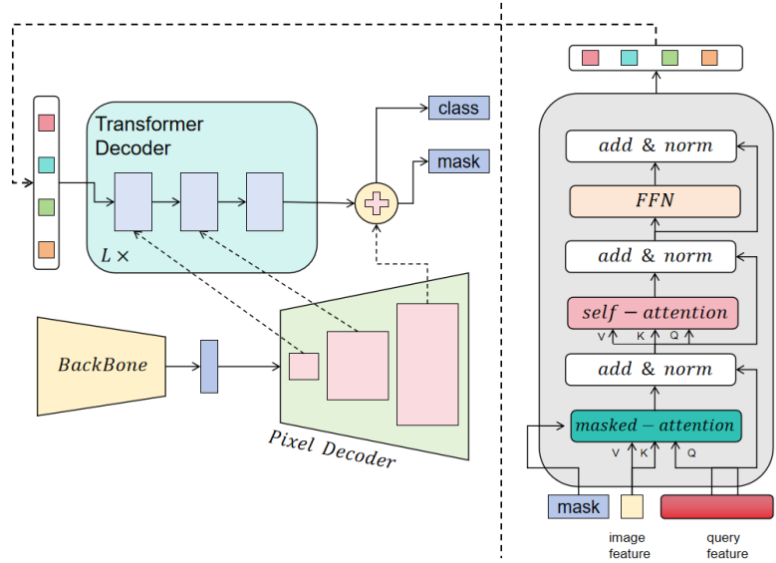

Figure 1. Mask2Former basic architecture

First, define the input image $I \in R^{H \times W \times 3}$, and extract multi-scale features $\{F_l\}_{l=1}^L$ through the backbone network, where $F_l \in R^{H_l \times W_l \times C_l}$ represents the feature map of the l-th layer. For each scale feature, we transform the feature through the Transformer encoder to obtain the feature representation after global perception:

$$Z_l = TransformerEncoder(F_l)$$

Among them, $Z_l$ represents the feature representation obtained after passing through the Transformer encoder.

In order to improve the expressiveness of boundaries, we propose the Boundary Enhanced Feature Bridging Module (BEFBM), which combines advanced boundary detection methodologies [11] with generative design insights [12]. At its core, BEFBM introduces a gradient-guided boundary enhancement loss that focuses on precisely delineating the contours of key regions, ensuring that subtle edges are not overlooked. Additionally, cross-scale feature fusion allows for the integration of high-resolution boundary cues with broader contextual information, enhancing the model's ability to distinguish between closely adjacent objects. By aligning these strategies with robust few-shot learning approaches [13], BEFBM demonstrates resilience even when training data is scarce, enabling effective boundary detection and refinement across diverse applications. This comprehensive framework ultimately bolsters both the precision and generalizability of boundary-related tasks in the broader scope of computer vision research, which is central to this paper's goals. First, we use the gradient operator to extract the edge information of the image:

$$E = \sqrt{(\frac{\partial I}{\partial x})^2 + (\frac{\partial I}{\partial y})^2}$$

Where $E \in R^{H \times W}$ represents the edge information of the input image. During the training process, we calculate the similarity between the feature map and the edge information so that the feature can learn more boundary details:

$$L_{edge} = \sum_{l=1}^{L} \| Z_l - E_l \|^2$$

Among them, $E_l$ represents downsampling the edge information to match the resolution of $Z_l$. This loss function ensures that the model can pay attention to the details of the target boundary while learning global features.

In addition, we introduce a feature bridging mechanism in BEFBM to enhance information interaction between features of different scales through the attention mechanism. Building on adaptive weighting strategies [14] and multi-scale attention approaches [15], this mechanism effectively captures both local details and broader context. Specifically, given two features and of different scales, we fuse them via an adaptive weighted approach, ensuring a balanced representation across varying feature hierarchies. Further guided by graph-based hierarchical insights [16], the fused features promote robust learning while maintaining flexibility in accommodating diverse and complex data distributions:

$$F_{bridge} = \alpha \cdot Z_i + (1 - \alpha) \cdot Z_j$$

Among them, $\alpha$ is calculated by attention weight:

$$\alpha = \sigma(W_1 \cdot GAP(Z_i) + W_2 \cdot GAP(Z_j))$$

Among them, $GAP(\cdot)$ represents global average pooling, $W_1$ and $W_2$ are learnable parameters, and $\sigma(\cdot)$ is the Sigmoid activation function. This mechanism ensures that features of different scales can be fused according to adaptive weights, thereby improving the expressiveness of cross-scale features.

In the final segmentation prediction stage, we employ a query-driven mask prediction mechanism, encompassing both category and mask generation. This design specifically focuses on integrating category-level information with pixel-level details, thereby refining the final segmentation accuracy. Building on cross-scale attention and multi-layer feature fusion principles [17], the model aligns target detection insights with semantic segmentation objectives, ensuring boundary precision and contextual consistency. Furthermore, the incorporation of knowledge graph-based fine-tuning [18] supports adaptive learning in complex, domain-specific scenarios, while advanced summarization methods [19] highlight the interpretative flexibility offered by a query-based architecture. Given a query set , the Transformer decoder updates it:

$$Q(t) = TransformerDecoder(Q^{(t-1)}, z)$$

Among them, $Q(t)$ represents the updated query vector after the tth round of decoding. Finally, we use the query vector to generate the semantic segmentation mask:

$$M_k = \sigma(W_3 \cdot Q_k)$$

Among them, $W_3$ is the learnable parameter and $M_k$ is the predicted mask of the k-th target.

In order to optimize the training process, we designed a comprehensive loss function, including classification loss $L_{cls}$, mask loss $L_{mask}$ and boundary enhancement loss $L_{edge}$.

$$L = \lambda_1 L_{cls} + \lambda_2 L_{mask} + \lambda_3 L_{edge}$$

Among them, $\lambda_1, \lambda_2, \lambda_3$ is a hyperparameter used to balance the impact of different loss terms. Finally, by introducing the boundary enhancement feature bridging module, Mask2Former can more effectively capture boundary information and improve the accuracy and robustness of semantic segmentation.

## IV. EXPERIMENT

### A. Datasets

This study employs the Cityscapes dataset to evaluate the proposed Mask2Former algorithm, which incorporates a boundary-enhancement feature bridging module for semantic segmentation. Cityscapes is a high-quality benchmark composed of high-resolution images (2048×1024) from 50 cities, encompassing diverse and complex urban traffic scenes. It provides fine-grained pixel-level and instance-level annotations for multiple categories (e.g., roads, pedestrians, vehicles, buildings), effectively capturing the real-world challenges relevant to autonomous driving and urban planning.

A central difficulty in Cityscapes is the dynamic complexity and uncertain boundaries among objects. Significant scale variations exist between distant pedestrians and nearby vehicles, making multi-scale feature learning more demanding. Furthermore, boundary ambiguity—such as the overlap between pedestrians and bicycles or vehicles and road surfaces—leads to misclassification, resulting in object adhesion or fragmented edges in traditional methods. In this work, the standard training set (2975 images), validation set (500 images), and test set (1525 images) of Cityscapes are used to thoroughly assess model performance. Due to the high resolution of the data, random cropping and scaling are applied for data augmentation during training, thereby reducing computational load and enhancing generalization. To further refine boundary detection, Sobel-based edge maps are integrated as supplementary supervision signals, strengthening the model's focus on target boundaries. These comprehensive experimental configurations validate the effectiveness of the proposed approach and establish a robust basis for subsequent model optimization.

## B. Experimental Results

To start, a comparative analysis between this model and other models is presented, along with the experimental results presented in Table 1.

Table 1. Experimental results

| Model | mIOU | mDICE | mRecall | FPS |
|-------|------|-------|---------|-----|
| FCN | 62.5 | 71.3 | 68.9 | 42 |
| Unet | 65.8 | 74.2 | 72.1 | 38 |
| Deeplabv3+ | 71.2 | 79.4 | 77.6 | 34 |
| SegFormer | 74.1 | 82.5 | 81.2 | 25 |
| Mask2Former | 78.5 | 85.7 | 84.3 | 28 |
| Ours | 81.3 | 88.1 | 86.9 | 24 |

The experimental results indicate that different semantic segmentation models vary significantly in mIOU, mDICE, and mRecall. Traditional CNN-based methods (e.g., FCN and UNet) show modest performance (mIOU of 62.5 and 65.8, respectively) and struggle with complex boundaries [20]. DeepLabv3+ achieves higher accuracy (mIOU=71.2, mDICE=79.4) by leveraging dilated convolutions and stronger context modeling, but its inference speed (34 FPS) lags behind FCN [21] and UNet [22]. Transformer-based approaches exhibit stronger feature representation. SegFormer attains mIOU=74.1 and mDICE=82.5, while Mask2Former further improves mIOU to 78.5, mDICE to 85.7, and inference speed to 28 FPS (compared to 25 FPS for SegFormer). Our proposed boundary-enhancement feature bridge module achieves the highest performance (mIOU=81.3, mDICE=88.1, mRecall=86.9), though its speed (24 FPS) is slightly lower than Mask2Former. Overall, it significantly improves boundary delineation and segmentation consistency. Figure 2 presents the training loss curve.

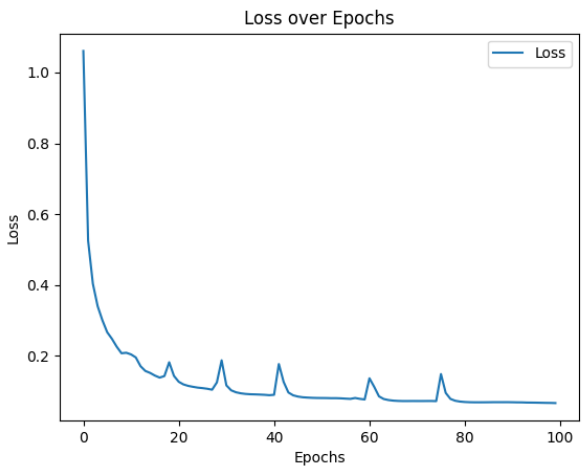

Figure 2. Loss function drop graph

This figure shows the trend of the loss function changing with epochs during training. From the curve, the loss value drops rapidly in the early stage of training, indicating that the model can quickly learn useful features and optimize parameters in the early stage. After about 20 epochs, the loss curve tends to be stable, but there is still some fluctuation, indicating that the model is still undergoing subtle adjustments. Overall, the loss value finally converges to a lower level, indicating that the model has successfully completed the training and has good convergence.

However, the fluctuation phenomenon in the curve may mean that the model has encountered training instability at some stages, such as too high learning rates or large batch gradient changes. This fluctuation may be related to the optimization strategy, such as using a larger learning rate or dynamic learning rate adjustment. In addition, if the fluctuation is large and there is no obvious convergence, it may mean that the model is at risk of underfitting or overfitting. Therefore, when further optimizing the model, you can consider adjusting the learning rate strategy, optimizing the regularization method, or adopting a more stable gradient update mechanism to improve the stability and generalization ability of the model.

Finally, an intuitive segmentation result is given, as shown in Figure 3.

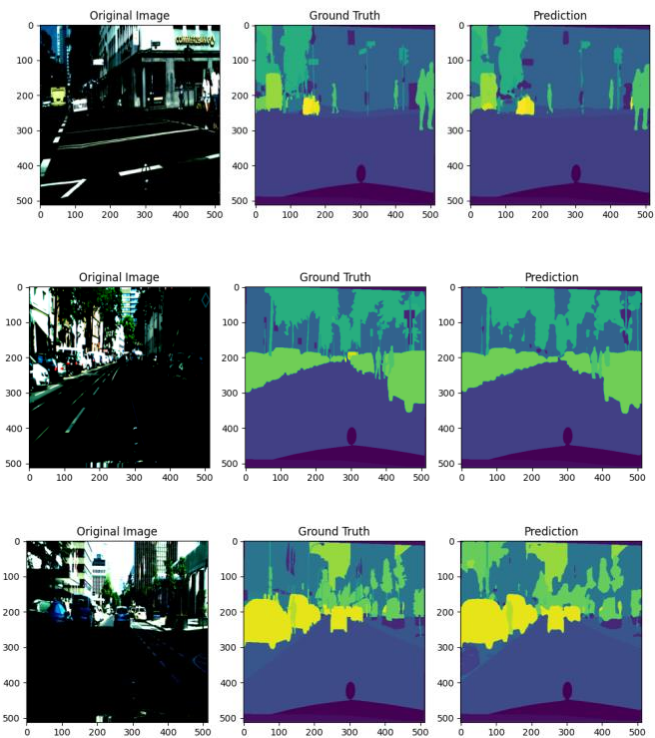

Figure 3. Semantic segmentation intuitive results

From the visualization results of semantic segmentation, the prediction results of the model have a high degree of similarity with the ground truth overall and can accurately identify target areas of different categories, such as roads, buildings, pedestrians, and vehicles. The segmentation results can better maintain the scene structure globally, especially since the segmentation effect of large areas such as roads and skies is relatively stable. However, in some details, such as pedestrian boundaries, tree outlines, and small target areas (such as vehicles), the prediction results still have a certain deviation, and there is a certain degree of misalignment with the real labels, indicating that the model still has room for improvement in the processing of boundary details.

In addition, from the comparison of multiple examples, the segmentation quality of the model has declined in complex backgrounds or poor lighting conditions (such as the first and second rows of images), especially in the recognition of shadow areas and distant targets. This may be due to the loss of boundary features during feature extraction or the distribution of samples in the data set causing the model to learn insufficiently in some categories. The direction of improvement can consider the use of boundary enhancement methods, such as adding additional boundary supervision losses or improving the cross-scale information interaction capabilities through feature bridging modules, so as to further optimize the segmentation accuracy of target contours and small target areas. Overall, the model has achieved high-quality semantic segmentation, but there is still room for improvement, especially in terms of boundary details and generalization capabilities for complex scenes.

## V. CONCLUSION

This study proposes a Mask2Former semantic segmentation algorithm based on the boundary enhancement feature module (BEFBM) to improve the accuracy and segmentation consistency of target boundaries. By introducing boundary information guidance mechanism and cross-scale feature fusion strategy in Transformer architecture, the experimental results of the model on the Cityscapes dataset show that compared with the existing semantic segmentation methods, the proposed method has achieved significant improvements in indicators such as mIOU, mDICE and mRecall. At the same time, the loss convergence curve and visualization results further prove that the proposed method can effectively improve the clarity of target boundaries and maintain strong robustness in complex urban scenes. Although the computational complexity is slightly increased, the overall inference speed remains within an acceptable range, ensuring the practical application value of the method.

Future research can further explore more efficient boundary enhancement strategies to reduce computational overhead and improve the generalization ability of the model in different scenarios. In addition, weakly supervised learning or self-supervised learning methods can be combined to reduce the dependence on high-quality labeled data, thereby improving the applicability of the model on large-scale unlabeled data. In addition, for different application scenarios, such as remote sensing image analysis or medical image segmentation, the feature bridging module can be adapted and optimized to further enhance the adaptability and scalability of the segmentation model.